\documentclass{article}

\PassOptionsToPackage{numbers, compress}{natbib}
    \usepackage[final]{neurips_2019}

\usepackage[utf8]{inputenc} % allow utf-8 input
\usepackage[T1]{fontenc}    % use 8-bit T1 fonts
\usepackage{hyperref}       % hyperlinks
\usepackage{url}            % simple URL typesetting
\usepackage{booktabs}       % professional-quality tables
\usepackage{amsfonts}       % blackboard math symbols
\usepackage{nicefrac}       % compact symbols for 1/2, etc.
\usepackage{microtype}      % microtypography
\usepackage{graphicx}
\usepackage{amsmath}
\usepackage[ruled,vlined]{algorithm2e}

\title{To Populate Is To Regulate}

\author{
  Nicole Fitzgerald \\
  Microsoft Research Montreal \\
  \texttt{nifitzge@microsoft.com} \\
}

\begin{document}

\maketitle

\begin{abstract}
In this work, we examine the effects of instantiating Lewis signaling games within a population of speaker and listener agents with the aim of producing a set of general and robust representations of unstructured pixel data. Preliminary experiments suggest that the set of representations associated with languages generated within a population outperform those generated between a single speaker-listener pair on this objective, making a case for the adoption of population-based approaches in emergent communication studies. Furthermore, post-hoc analysis reveals that population-based learning induces a number of novel factors to the conventional emergent communication setup, inviting a wide range of future research questions regarding communication dynamics and the flow of information within them.

\end{abstract}

\section{Introduction}

Lewis signaling games~\cite{lewis1969convention} have found extensive use in linguistic and cognitive studies in the context of language evolution~\cite{briscoe2002linguistic,cangelosi2002computer,steels2003social}. More recently, however, they have seen widespread adoption in computational studies concerning the emergence of communication protocols between neural agents~\cite{lazaridou2018emergence,bouchacourt2018agents}, spawning a wave of contemporary research - from empirical investigation of the structural properties and task-oriented effectiveness \cite{cao2018emergent, bouchacourt2019miss} of the emerged languages to the formulation of robust methodology within which to frame that investigation~\cite{lowe2019pitfalls}.

While the bulk of existing research has focused its attention on the nature and nuances of these emerged languages, that of the representations which underlie and inform them have, by and large, escaped comprehensive scrutiny. The interplay between natural language and world, or the representation thereof, has long been the subject of immense debate - we refer here to~\cite{wittgenstein2013tractatus,frege1948sense,quine2013word, davidson1984very} as a selection of foundational works which have greatly shaped the development of recent literature. 

This work looks to investigate the extent to which the sets of representations associated with emergent languages are, in fact, general and robust. Aligning with the motivation of \cite{tielemanshaping}, we conjecture that emergent languages generated by a single pair of speakers may be prone to capturing non-general and highly particular representations. Drawing inspiration from the evolution of natural language, we proffer the additional conjecture that emerging languages amongst a population of agents is more likely to induce a more desirable set of underlying features within the representation space constructed by individual agents.

To this end, this work extends the typical two-player game setup~\cite{lazaridou2018emergence} to a multi-player variant in which agents situated within a population play several two-player games, each with a unique set of partners sampled from within that same population. We propose this particular setup as a method for examining the extent to which communicating with a diversity of agents affects the formation of pre-linguistic representations. 

Notably, the works of \cite{lowelearning} and \cite{tielemanshaping} have both previously introduced the concept of population-based emergent to the literature. We distinguish this work from the former in that \cite{lowelearning} seeks to leverage the diversity of languages emerging in various populations as a suite of meta-learning tasks, rather than investigating the properties of those languages directly. We distinguish this work from the latter in that \cite{tielemanshaping} seeks to learn an auto-encoding task rather than a traditional signaling game and trains encoder-decoder pairs iteratively over single training steps, rather than over a sequence of steps. Furthermore, we distinguish this work from the both of them in that we pass speaker messages through a discrete, i.e. non-differentiable channel, more aptly modeling the conventional communication method of natural language. 

We evaluate the generality of the representations acquired in both the conventional signaling game formulation and those acquired in our population-based formulation by investigating how quickly and how well novel speaker-listener pairs learn to communicate to solve problems unseen during training. Conforming to expectation, we find that the population-based method indeed exhibits superior performance on these objectives. We refer to Section \ref{sec:results} for more detailed analysis.

\section{Experimental Set-Up}
\subsection{Lewis Signaling Games}
\label{subsec: Lewis Signaling Games}
In the conventional setup, a \emph{speaker} agent is presented with a target image $x_t$ while a corresponding \emph{listener} agent is presented with a set of candidate images $X = \{x_1,...,x_n\}, x_t \in X$ which contains the target image and $|X| - 1$ distractor images. During gameplay, the speaker selects symbols from a vocabulary $W$ to construct a message $m = \{w_1,...,w_m\}, w_i \in W$ which serves as a description of the target image. Given $m$, the listener agent must then identify the target from its set of candidates. \emph{Communicative success} is defined as the correct identification of the target image by the listening agent \cite{lazaridou2018emergence}. In the following subsection, we will refer to a single played game as an LSG.

\subsection{Population-Based Learning}
Prior to training, we initialize a \emph{population} of $n$ agents. In this setting, we iterate over an \emph{outer loop}, which is comprised of sampling and subsequently training, a single speaker-listener pair on a succession of games, until some stopping criterion - either a fixed number of pair steps or a threshold value of communicative success -  is reached. The training of the single speaker-listener pair will be considered the \emph{inner loop} of our training paradigm.

From a general standpoint, experiments conducted within this framework enjoy three principal degrees of freedom: (1) $n$, denoting the size of the population, and (2) $k$, denoting the number of partners with which a given agent trains, and (3) $s$, denoting the number of steps for which a given pair trains. 

\begin{algorithm}[H]
\SetAlgoLined
\textbf{Require} $n$: the population size\;
\textbf{Require} $k$: the number of partners per agent\;
\textbf{Require} $s$ the number of train steps per pair\;
\smallskip
\textbf{Compute} \emph{total\_pairs} $\leftarrow{k * n}$\;
\smallskip
\textbf{Initialize} speakers $S = \{s_1, s_2,...,s_{\frac{1}{2}n}\}$\;
\textbf{Initialize} listeners $L = \{l_1, l_2,...,l_{\frac{1}{2}n}\}$\;
\smallskip
 \For{pair \normalfont{in} \emph{total\_pairs}}{
  sample $s_i \in S$\;
  sample $l_j \in L$\;
  \For{step \normalfont{in} \emph{s}}{
  \emph{t'} $\leftarrow{LSG(s_i, l_j)}$ // compute pair prediction on a single game\;
  $L_{pair}$ $\leftarrow{L(t, t')}$ // compute loss given target and prediction\;
  optimize $s_i$ and $l_j$ given $L_{pair}$
   }{
   return $s_i$ to $S$\;
   return $l_j$ to $L$\;
  }
 }
\caption{Population-Based Learning (PBL)}
\end{algorithm}

In this particular work, we aim to train the population uniformly, thus we construct a given population to contain an equal ratio of speakers and listeners and we fix the number of training partners for all agents such that $k=4$. Over our suite of experiments, we vary the population size $n \in N=\{2, 6, 10\}$, where the $n = 2$ setting aims to model the conventional two-player setup as described in Section \ref{subsec: Lewis Signaling Games} . For each value of $n$, we initialize two disjoint populations of that size, a detail whose purpose will be made clear in Subsection \ref{subsec:eval}. Each population contains an equal number of speakers and listeners. We emphasize here that each population is trained strictly independently of all others. Each pair trains together for $s = 1 024 000$ steps. We refer to Appendix \ref{subsec: sampling} for the details of our sampling procedure.

Though this work optimizes only the performance of individual speaker-listener pairs, one can imagine a variant in which we optimize the performance of a population as a whole instead of or in addition to pair-based optimization.

\subsection{Evaluation}
\label{subsec:eval}
The primary concern of this work is to investigate the extent to which population size affects the generality and robustness of the representations associated with agents' communicative policies. We define \emph{generality} as the ability of a representation space to aptly interpret novel input and \emph{robustness} the ability to do so for any given speaker-listener pair. As such, we think it interesting to frame our evaluation process as an attempt to answer the following question: given the representations learned during training, how quickly and to what extent can a novel speaker-listener pair learn to communicate about previously unseen data?

To this end, the evaluation process is as follows: for each value of $n$, we sample a speaker-listener pair such that the speaker and the listener originate from different populations. We freeze the parameters associated with each agent's image encoder, namely those belonging to the convolutional layer and subsequent two-layer feed-forward network, and randomly initialize the LSTMs constituting the speaker and listener policies respectively. The speaker and listener policies are then optimized in the setting of a conventional two-player signaling game, which is constructed using a set of held-out test images drawn from the same distribution as the training set. 

To induce further novelty, the set of test games contains a greater number of distractor images than seen during training. We allow each pair to train for 1 024 000 steps. We present and analyze our results in Section \ref{sec:results} of this paper.

\section{Implementation Details}
\subsection{Agents}
\label{subsec:agents}
We adopt the speaker and listener architectures employed in~\cite{lazaridou2018emergence}
with the addition of a pre-trained vision module $f(\theta, x)$, implemented by a single-layer convolutional network followed by a two-layer feed-forward network to encode the images. As noted above, all parameters belonging to this module are frozen during the evaluation phase. 

The speaker encodes the target image $x_t$ into a dense vector representation $u_t$ via its image encoder $f^{S}(\theta^{S}_{f}, x_t)$. The speaker constructs a fixed-length message \textbf{m} of at most length $L$ by sampling a token from a vocabulary V of discrete tokens at each time step according to a recurrent policy $\pi^{S}$ generated by a single layer LSTM~\cite{hochreiter1997long} $h^{S}(\theta_{h}^{S}, u_t)$.

The listener is implemented in a similar fashion, wherein each image in its set of candidates is encoded via the encoder $f^L(\theta^{L}_{f}, x_i)$ to form a set of image representations $U = \{u_{i} = f^L(\theta^{L}_{f}, x_i) | x_i \in X\}$. The message \textbf{m} is encoded via a single-layer LSTM $h^L(\theta_{h}^L, z)$, where $z$ is the embedded vector representation of \textbf{m}. The listener selects an image $t'$ by sampling from a Gibbs distribution generated via the dot product of $z$ and all encoded images $u \in U$.

We refer to Appendix \ref{subsec: training} for the full details of the the agent architectures.

\subsection{Learning}
We pre-train the image encoders on the set of training images. In the pre-training phase, we train a vision module comprised of a single convolution layer and subsequent two-layer linear classifier to predict both the color and the position of the shape contained in each image. The color and position of a given shape are denoted by a two-dimensional label vector $y \in Y$  as specified in Section \ref{subsec:agents}. 

During pre-training, the classifier is trained to minimize the the categorical cross-entropy loss over the the set of labels $Y$. In the training phase, we discard this classifier in favour of a randomly initialized two-layer MLP, keeping only the convolutional layer in order to minimize the amount of structural bias encoded by the vision module while maintaining a nominal visual prior.  

At training time, the full model parameters of both the speaker and listener are optimized in tandem over a batch of games. The objective function optimized by speaker-listener pair may be denoted by,
\begin{equation}
% agent loss fn
L(\theta_{f}^S, \theta_{h}^S, \theta_{f}^L, \theta_{h}^L) = ((R(t')-b) \cdot (\sum_{l=1}^L \log p_{{\pi}^{S}}(m^{l}_{t}|m^{<l}_t, u_{t}) + \log p_{{\pi}^{L}}(t'|z, U)) - H_S
\end{equation}
where $b$ is a baseline variance reduction term that we simply set to $b = {1 \over N} \sum r(\tau)$, $R(t')$ is the reward, which is $1$ if the predicted target is correct (i.e. $t=t'$) and 0 otherwise, and $z$ is the encoding of the message \textbf{m} computed by the listener LSTM. The entropy term $H_S$ corresponding to the entropy of the speaker's policy is a regularization term added to encourage exploration \cite{mnih2016asynchronous}.

During evaluation, we freeze all parameters associated with the vision module, i.e. the speaker and listener convolutional layers, and train only those parameters associated with the communication protocol, i.e. the speaker and listener LSTMS, in addition to the listener's pointing module.  Hence, we optimize the function,

\begin{equation}
% agent loss fn
L(\theta_{h}^S, \theta_{h}^L) = ((R(t')-b) \cdot (\sum_{l=1}^L \log p_{{\pi}^{S}}(m^{l}_{t}|m^{<l}_t, u_{t}) + \log p_{{\pi}^{L}}(t'|z, U)) - H_S
\end{equation}

Given the discrete nature of the messages, we estimate model parameters via the REINFORCE update rule \cite{williams1992simple}. As previously noted, the parameters $\theta_{h}^S$ and $\theta_{h}^L$ are randomly initialized at the beginning of the evaluation phase.

\section{Results \& Discussion}
\label{sec:results}
Fig. 1, below, illustrates the averaged performance of 4 randomly selected speaker-listener pairs on the evaluation task described in section 2.3. Evaluation was performed over a set of candidate images of size $|X| = 5$, thus the random baseline in this setting analytically yields a mean reward of 0.2. 

\begin{figure}[htp]
    \centering
    \includegraphics[width=12cm]{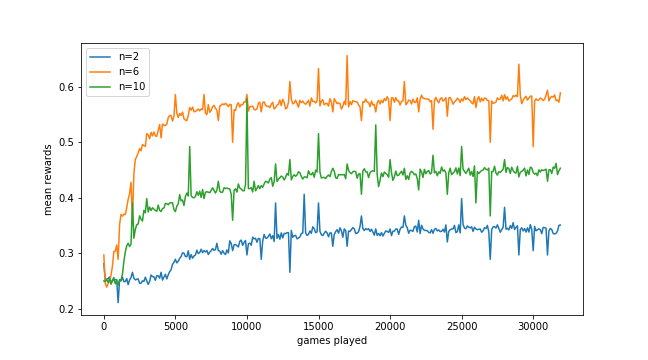}
    \caption{Learning curves of speaker-listener pairs on the designated evaluation task.}
    \label{fig:results}
\end{figure}

Results depicted in figure \ref{fig:results} suggest the population-based methods outperform agents trained in the conventional two-player setup, both in terms of sample efficiency and overall accuracy. This confirms the central tenet of the original hypothesis, namely that languages, and by extension their underlying representations, emerging in the context of a population are subject to an implicit regularizing effect attributed to the size and structure of that population. We note, however, that agents trained within the population of 6 agents significantly outperform agents trained within the larger population of 10.

% ugh
Formally, we may conceive of our network of pairings within each $n$-sized population as a random $k$-regular bipartite graph containing $n$ nodes. Furthermore, we can assign some weight $w$ to each edge, denoting number of times a given speaker has been paired with a given listener. Given some fixed $k$, it is generally assumed that the measure of sparsity within the graph grows as we scale $n$. As such, the topology of the graph evolves dramatically with the increase in $n$, exhibiting higher probability of potentially critical phenomena, such an increase of the minimal path between a given speaker-listener pair; the emergence of minimally-weighted edges, anti-edges and islands; and increasingly small cut-sets, each of which may affect language convergence to varying degrees. 

The rich expanse of existing literature in random graph theory leaves us well-equipped to develop a formal framework under which to better analyze the dynamics of language emergence and convergence conditions within population-based methods. Though preliminary, our results demonstrate the value of this approach and beg a multitude of subsequent questions. How does the population topology affect the both the structural properties of emergent languages and their underlying representations? Is there an optimal population topology to promote the emergence of non-natural language? Anthropologically, does this mimic the population topology and spread of natural languages? With this in mind, we believe that population-based approaches to emergent communication studies hold immense potential for a vibrant wave of subsequent research. We are excited to further our work in this direction and hope it provides the grounds for much fruitful discussion.

\subsection*{Acknowledgments}

Extensive thanks is owed to many of my colleagues at MSR Montreal, in particular Kaheer Suleman and Alessandro Sordoni, for their support and enormously constructive feedback. Additional thanks to Tavian Barnes, who knows a lot about graphs.

\bibliography{main}
\bibliographystyle{plain}

\appendix

\section{Appendix}
\subsection{Sampling Procedure}
\label{subsec: sampling}
\paragraph{Training} Within a given population, we assign to each speaker a unique id, $s_i \in (0, n\_speakers]$,  $s_i \in \mathbb{Z}$ and generate a list $S$ such that each unique $s_i$ has $k$ occurrences in $S$. The list $L$ of listener ids is constructed in a similar fashion. The orderings of both $S$ and $L$ are randomly shuffled and pairings are constructed by iterating through the two lists concurrently.

\paragraph{Evaluation} For each value of $n$, we construct a list of speaker ids $S$ and listener ids $L$ as above. For some value $n\_test\_pairs$, we sample that many ids from each $S$ and $L$, without replacement. The are randomly shuffled and pairing are constructed via concurrent iteration.

\subsection{Data}
\paragraph{Training} As in \cite{lazaridou2018emergence}, we construct our set of training games from a set of 4000 synthetic images of geometric shapes, generated by the Mujoco physics engine. A single game is generated by randomly sampling a set of $X$ images from the dataset and subsequently sampling a single image $x_t \in X$ as the target image. 

\begin{figure}[htp]
    \centering
    \includegraphics[width=3cm]{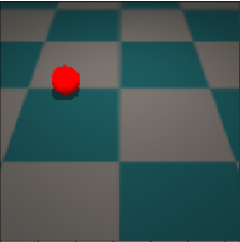}
    \includegraphics[width=3cm]{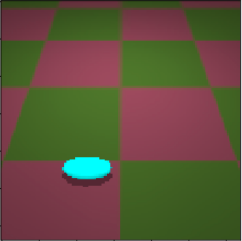}
    \includegraphics[width=3cm]{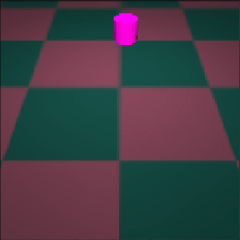}
    \caption{Training examples}
    \label{fig:results}
\end{figure}

\paragraph{Pre-Training} Associated with each image in our dataset is a 2-dimensional vector $y = [y_1, y_2]$ s.t. $y_1, y_2 \in \mathbb{Z}$ where $y_1$ denotes the color of the shape and $y_2$ denotes its position in the image. The color attribute $y_1$ takes on one of 8 possible values, while the position vector takes on one of 5 possible values.

\paragraph{Evaluation} The test set is similarly constructed over a collection of 1000 images from the same source.  

\subsection{Training Details}
\label{subsec: training}
We instantiated our experiments with the following hyper-parameters:

The convolutional layer has $n\_in\_channels = 3$ and $n\_out\_channels = 20$, with a kernel size of 5 and stride of 1. The subsequent MLP has a hidden size and output size of 50. 

The size of the vocabulary, i.e. $|V| = 20$ and the message length $L$ is at most 5. We set the dimension of the embedding matrix used to embed the message tokens to be size 32. The dimensionality of the speaker and listener LSTM hidden states is 64. 

We multiply the speaker entropy term by a coefficient $\alpha = 0.1-|(R(t') - b| \cdot 0.1$ while $speaker\_steps < 1 000 000$ and $0.01$ otherwise.

We train the agents with a set of 4 candidate images, i.e. $|X| = 4$, and evaluate on a set of 5 candidate images, i.e. $|X| = 5$.

\end{document}